\documentclass[manuscript, screen]{acmart}
\usepackage{booktabs} 

\usepackage[ruled]{algorithm2e} 

\usepackage{todonotes}
\usepackage{subcaption}

\newcommand\blfootnote[1]{%
  \begingroup
  \renewcommand\thefootnote{}\footnote{#1}%
  \addtocounter{footnote}{-1}%
  \endgroup
}

\acmJournal{JDIQ}
\acmVolume{11} 
\acmNumber{3}
\acmArticle{12}
\acmYear{2019}
\acmMonth{5}

\setcopyright{acmcopyright}

\acmDOI{3331015.3297722}

\newcommand{\blue}[1]{\textcolor{blue}{#1}}

\newcommand{\Ni}{({\em i})~}
\newcommand{\Nii}{({\em ii})~}
\newcommand{\Niii}{({\em iii})~}

\newcommand{\good}{\textsc{Good}\,}
\newcommand{\bad}{\textsc{Bad}\,}
\newcommand{\potential}{\textsc{Potentially Useful}\,}

\newcommand{\contentLingf}{Linguistic bias, subjectivity and sentiment}

\newcommand{\contentCredibility}{Credibility}

\newcommand{\forumRankSameUser}{Support from the current thread}

\newcommand{\forumIRQLsplit}{Support from all of Qatar Living}

\newcommand{\forumHighQualPosts}{Support from high-quality posts in Qatar Living}

\newcommand{\extIRFBingWebpageQataronlySplitted}{Support from the Web}
\usepackage{color}
\newcommand{\mCBP}{CB online~}
\newcommand{\mRANLPsingle}{singletonG~}
\newcommand{\mRANLP}{\emph{any~}}
\newcommand{\mCBfeats}{singleton CB~}

\newcommand{\mA}{{\em singleton}~}
\newcommand{\mB}{{\em any}~}
\newcommand{\mC}{{\em singleton+any}~}
\newcommand{\mD}{{\em multi}~}
\newcommand{\mE}{{\em multi+any}~} 

\begin{document}

\author{Pepa Atanasova}
\affiliation{%
\institution{Sofia University ``St. Kliment Ohridski''}
  \streetaddress{5, James Bourchier Blvd., 1164}
  \city{Sofia}
  \country{Bulgaria}
}
\email{pepa.k.gencheva@gmail.com}
\author{Preslav Nakov}
\affiliation{%
\institution{Qatar Computing Research Institute, HBKU}
  \streetaddress{HBKU Research Complex B1, P.O. Box 5825}
  \city{Doha}
  \country{Qatar}
}
\email{pnakov@qf.org.qa}
\author{Llu\'is M\`arquez$^*$}
\affiliation{%
\institution{Amazon}
  \streetaddress{}
  \city{Barcelona}
  \country{Spain}
}
\email{lluismv@amazon.com}
\author{Alberto Barr\'on-Cede\~no}
\affiliation{%
\institution{Qatar Computing Research Institute, HBKU}
  \streetaddress{HBKU Research Complex B1, P.O. Box 5825}
  \city{Doha}
  \country{Qatar}
}
\email{albarron@[hbku.edu.qa|gmail.com]}
\author{Georgi Karadzhov}
\affiliation{%
\institution{Sofia University ``St. Kliment Ohridski''}
  \streetaddress{5, James Bourchier Blvd., 1164}
  \city{Sofia}
  \country{Bulgaria}
}
\email{georgi.m.karadjov@gmail.com}
\author{Tsvetomila Mihaylova}
\affiliation{%
\institution{Sofia University ``St. Kliment Ohridski''}
  \streetaddress{5, James Bourchier Blvd., 11641}
  \city{Sofia}
  \country{Bulgaria}
}
\email{tsvetomila.mihaylova@gmail.com}
\author{Mitra Mohtarami}
\affiliation{%
\institution{Massachusetts Institute of Technology}
  \streetaddress{MIT Computer Science and Artificial Intelligence Laboratory}
  \city{Cambridge}
  \state{MA}
  \country{USA}
}
\email{mitra@csail.mit.edu}
\author{James Glass}
\affiliation{%
\institution{Massachusetts Institute of Technology}
  \streetaddress{MIT Computer Science and Artificial Intelligence Laboratory}
  \city{Cambridge}
  \state{MA}
  \country{USA}
}
\email{glass@mit.edu}

\title{Automatic Fact-Checking Using Context and Discourse Information}

\begin{abstract}
We study the problem of automatic fact-checking, paying special attention to the impact of contextual and discourse information. We address two related tasks: (\emph{i})~detecting check-worthy claims, and (\emph{ii})~fact-checking claims.
We develop supervised systems based on neural networks, kernel-based support vector machines, and combinations thereof, which make use of rich input representations in terms of discourse cues
and contextual features. For the check-worthiness estimation task, we focus on political debates, and we model the target claim in the context of the full intervention of a participant and the previous and the following turns in the debate, taking into account contextual meta information. For the fact-checking task, we focus on answer verification in a community forum, and we model the veracity of the answer with respect to the entire question--answer thread in which it occurs as well as with respect to other related posts from the entire forum. 
We develop annotated datasets for both tasks and we run extensive experimental evaluation, confirming that both types of information ---but especially contextual features--- play an important role.
\end{abstract}

%
%


%
%


\blfootnote{$^*$ Work conducted while this author was at QCRI, HBKU.}
\maketitle

\renewcommand{\shortauthors}{Atanasova et al.}


\section{Introduction}
\label{sec:intro}


Recent years have seen the proliferation of deceptive information online. With the increasing necessity to validate information from the Internet, \emph{automatic fact-checking} has emerged as an important research topic. Fact-checking is at the core of multiple applications, e.g.,  discovery of fake news \cite{Lazer1094}, rumor detection in social media \cite{Vosoughi1146}, information verification in question answering systems \cite{AAAI2018:factchecking}, detection of information manipulation agents \cite{Chen:2013:BIW:2492517.2492637,Mihaylov2015ExposingPO,SeminarUsers2017}, and assistive technologies for investigative journalism \cite{Hassan:15}. It touches many aspects, such as credibility of users and sources, information veracity, information verification, and linguistic aspects of deceptive language. 
There has been work on automatic claim identification~\cite{Hassan:15,Hassan2016ComparingAF}, and also on checking the factuality/credibility of a claim, of a news article, or of an information source~\cite{Castillo:2011:ICT:1963405.1963500,Ba:2016:VERA,PlosONE:2016,ma2016detecting,Hardalov2016,RANLP2017:clickbait,RANLP2017:factchecking:external,RANLP2017:credibility:trolls,D17-1317}. In general, previous work has not paid much attention to explicitly modeling contextual information and linguistic properties of the discourse in order to identify and verify claims, with some rare recent exceptions~\cite{popat2017,pepaRANLP:17}.

In this article, we focus on studying the role of \emph{contextual information} and \emph{discourse}, which provide important information that is typically not included in the usual feature sets, which are mostly based on properties of the target claim, and its similarity to a set of validation documents or snippets. In particular, we focus on the following tasks:

\paragraph{Check-worthy claim identification}
We address the automatic identification of claims in political debates which a journalist should fact-check. In this case, the text is dialog-style: with long turns by the candidates and orchestrated by a moderator around particular topics. Journalists had to challenge the veracity of claims in the 2016 US presidential campaign, and this was particularly challenging during the debates as a journalist had to prioritize which claims to fact-check first. Thus, we developed a model that ranks the claims by their check-worthiness.

\paragraph{Answers fact-checking}
We address the automatic verification of answers in community-driven Web forums (e.g.,~Quora, StackOverflow). The text is thread-style, but is subject to potential dialogues: a user posts a question and others post potential answers. That is, the answers are verified in the context of discussion threads in a forum and are also interpreted in the context of an initial question.
Here we deal with social media content. The text is noisier
%
and the information being shared is not always factual; mainly due to misunderstanding, ignorance, or maliciousness of the responder. 

We run extensive experiments for both tasks by training and applying classifiers based on neural networks, kernel-based support vector machines, and combinations thereof. The results confirm that the contextual and the discourse information are crucial to boost the models and to achieve state-of-the-art results for both tasks.%
\footnote{We make available the datasets and source code for both tasks: \url{https://github.com/pgencheva/claim-rank} and 
\url{https://github.com/qcri/QLFactChecking}}
In the former task, using context yields 4.2 MAP points of absolute improvement, while using discourse information adds 1.5 MAP absolute points; in the latter task, considering the discourse and the contextual information improves the performance by a total of 4.5 MAP absolute points.

The rest of this article is organized as follows: Section~\ref{sec:debates} describes our supervised approach to predicting the check-worthiness of text fragments with focus on political debates. Section~\ref{sec:cqa} presents our approach to verifying the factuality of the answers in a community question answering forum. 
Section~\ref{sec:discussion} provides a more qualitative analysis of the outcome of all our experiments. 
Section~\ref{sec:related} discusses related work.
Finally, Section~\ref{sec:conclusions} presents the conclusions and the lessons learned, and further outlines some possible directions for future research.


\section{Claim Identification}
\label{sec:debates}

In this section, we focus on the problem of automatically identifying which claims in a given document are most check-worthy and thus should be prioritized for fact-checking. We focus on how contextual and discourse information can help in this task. We further study how to learn from multiple sources simultaneously (e.g., PolitiFact, FactCheck, ABC), with the objective of mimicking the selection strategies of one particular target source;  we do this in a multi-task learning setup.

\subsection{Data}

We used the CW-USPD-2016 dataset, which is centered around political debates~\cite{pepaRANLP:17}. It contains four transcripts of the 2016 US Presidential election debates: one vice-presidential and three presidential. Each debate is annotated at the sentence level as \emph{check-worthy} or not, but the sentences are kept in the context of the full debate, including metadata about the speaker, speaker turns, and system messages about the public reaction.
The annotations were derived using publicly-available manual analysis of these debates by nine reputable fact-checking sources, shown in Table~\ref{table:per:medium}. 
This analysis was converted into a binary annotation: whether a particular sentence was annotated for factuality by a given source.
Whenever one or more annotations were about part of a sentence, the entire sentence was selected, and when an annotation spanned over multiple sentences, each of them was selected.
The dataset with the four debates contains 5,415 sentences, out of which 880 are positive examples (i.e.,~selected for fact-checking by at least one of the sources). Table~\ref{table:exampleA} presents an excerpt of this corpus.

Note that the investigative journalists did not select the check-worthy claims in isolation, ignoring the context. Our analysis shows that these include claims that were highly disputed during the debate, that were relevant to the topic introduced by the moderator, etc. We will make use of these contextual dependencies below.

\begin{table}
\small
\centering
\begin{tabular}{lrrrrr}
\hline
  \bf Medium & \bf 1st & \bf 2nd & \bf VP & \bf 3rd & \bf Total\\ 
 \hline
  ABC News& 35 & 50 & 29 & 28 & 142 \\
  Chicago Tribune & 30 & 29 & 31 & 38 & 128\\
  CNN & 46 & 30 & 37 & 60 & 173 \\
  FactCheck.org & 15 & 45 & 47 & 60 & 167 \\
  NPR & 99 & 92 & 91 & 89 & 371 \\
  PolitiFact & 74 & 62 & 60 & 57 & 253 \\
  The Guardian & 27 & 39 & 54 & 72 & 192 \\
  The New York Times & 26 & 25 & 46 & 52 & 149 \\
  The Washington Post & 26 & 19 & 33 & 17	 & 95 \\
 \hline
  \bf Total annotations & 378 & 391 & 428 & 473 &  1,670 \\
  \bf 
  Annotated sentences & 218 & 235 & 183 & 244 & 880 \\
  \hline
\end{tabular}
\caption{Number of annotations in each medium for the 1st, 2nd and 3rd presidential and the vice-presidential debates. The last row shows the number of annotated sentences that become the positive examples in the CW-USPD-2016 dataset.}
\label{table:per:medium}
\vspace{-1em}
\end{table}

\begin{table}[t]
\small
\centering
\begin{tabular}{l@{\hspace{1mm}}	p{63mm}c@{\hspace{1mm}}c@{\hspace{1mm}}c@{\hspace{1mm}}
c@{\hspace{1mm}}c@{\hspace{1mm}}c@{\hspace{1mm}}
c@{\hspace{1mm}}c@{\hspace{1mm}}c@{\hspace{1mm}}c@{\hspace{1mm}}c@{\hspace{1mm}}c}
\hline
\bf Speaker	& \bf Text	& \multicolumn{9}{c}{\bf Annotation Sources} \\
 &  & CT & ABC & CNN & WP & NPR & PF & TG & NYT & FC & \bf All & \bf Check?\\ 
\hline
Clinton: &	So we're now on the precipice of having a potentially much better economy, but the last thing we need to do is to go back to the policies that failed us in the first place. & 0 &	0 &	0 &	0 &	0 &	0 &	0 &	0 &	0 & 0 & No \\

\blue{Clinton:} &	\blue{Independent experts have looked at what I've proposed and looked at what Donald's proposed, and basically they've said this, that if his tax plan, which would blow up the debt by over \$5 trillion and would in some instances disadvantage middle-class families compared to the wealthy, were to go into effect, we would lose 3.5 million jobs and maybe have another recession.} &	\blue 1 &	\blue 1 &	\blue 0 &	\blue 0 &	\blue 1 &	\blue 1 &	\blue 0 &	\blue 1 &	\blue 1 & \blue 6 & \blue{Yes} \\

\blue{Clinton:} &	\blue{They've looked at my plans and they've said, OK, if we can do this, and I intend to get it done, we will have 10 million more new jobs, because we will be making investments where we can grow the economy.} &	\blue 1 &	\blue 0 &	\blue 0 &	\blue 0 &	\blue 0 &	\blue 0 &	\blue 0 &	\blue 0 &	\blue 0 &	\blue 1 & \blue{Yes}\\

Clinton: &	Take clean energy. &	0 &	0 &	0 &	0 &	0 &	0 &	0 &	0 &	0 & 0 & No\\

Clinton: &	Some country is going to be the clean- energy superpower of the 21st century. &	0	&	0 &	0 &	0 &	0 &	0 &	0 &	0 &	0 &	0 & No\\

\blue{Clinton:} &	\blue{Donald thinks that climate change is a hoax perpetrated by the Chinese.} &	\blue 1 &	\blue 1 &	\blue 1 &	\blue 1 &	\blue 0 &	\blue 0 &	\blue 1 &	\blue 0 &	\blue 1 &	\blue 6 & \blue{Yes}\\

Clinton: &	I think it's real. &	0 &	0 &	0 &	0 &	0 &	0 &	0 &	0 &	0 &	0 & No\\

\blue{Trump:} &	\blue{I did not.}
& 	\blue 1 &	\blue 1 &	\blue 0 &\blue 1 &	\blue 1 &	\blue 1 &	\blue 0 &	\blue 0 &	\blue 0 &	\blue 5 & \blue{Yes}\\

\hline
 \end{tabular}
\caption{Excerpt from the transcript of the first US Presidential Debate in 2016, annotated by nine sources: Chicago Tribune (CT), ABC News, CNN, Washington Post (WP), NPR, PolitiFact (PF), The Guardian (TG), The New York Times (NYT) and Factcheck.org (FC). Whether the media fact-checked the claim or not is indicated by a 1 or 0, respectively. The total number of sources that annotated an example is shown in column ``All''. Column ``Check?'' indicates the class label, i.e., whether the example is check-worthy or not.  The positive examples are also highlighted in blue.}
\label{table:exampleA}
\vspace{-2em}
\end{table}

\subsection{Modeling Context and Discourse}
\label{check_worthy_Modeling_Context}
We developed a rich input representation in order to model and to predict the \emph{check-worthiness} of a sentence. In particular, we included a variety of contextual and discourse-based features. They characterize the sentence in the context of the full~\emph{segment} by the same speaker, sometimes also looking at the previous and the following segments. We define a \emph{segment} as a maximal set of consecutive sentences by the same speaker, without intervention by another speaker or the moderator, i.e., a \emph{turn}. We start by describing these context-based features, which are the focus of attention of this work.

\subsubsection{Position (\emph{3 features})}  A sentence on the boundaries of a speaker's segment could contain a reaction to another statement or could provoke a reaction, which in turn could signal a check-worthy claim.
Thus, we added information about the position of the target sentence in its segment: whether it is first/last, as well as its reciprocal rank in the list of sentences in that segment. 

\subsubsection{Segment sizes (\emph{3 features})} The size of the segment belonging to one speaker might indicate whether the target sentence is part of a long speech, makes a short comment or is in the middle of a discussion with lots of interruptions. The size of the previous and of the next segments is also important in modeling the dialogue flow. Thus, we include three features with the size of the previous, the current, and the next segments. 

\subsubsection{Metadata (\emph{8 features})} Check-worthy claims often contain accusations about the opponents, as the example below shows (from the 2nd presidential debate):\medskip

{\begin{tabular}{rp{126mm}}
Trump:	& \textbf{Hillary Clinton} attacked those same women and attacked them viciously.	\\& ...\\
 Clinton:	&  They're doing it to try to influence the election for \textbf{Donald Trump}.\\\smallskip
\end{tabular}}

\noindent Thus, we use a feature that indicates whether the target sentence mentions the name of the opponent, whether the speaker is the moderator, and also who is speaking (3~features). We further use three binary features, indicating whether the target sentence is followed by 
a system message: 
\emph{applause}, \emph{laugh}, or \emph{cross-talk}.\medskip



\subsubsection{Topics (\emph{303 features})}
Some topics are more likely to be associated with check-worthy claims, and thus we have features modeling the topics in the target sentence as well as in the surrounding context.
We trained a Latent Dirichlet Allocation (LDA) topic model~\cite{blei2003latent} on all political speeches and debates in \emph{The American Presidency Project}\footnote{\url{http://www.presidency.ucsb.edu/debates.php}} using all US presidential debates in the 2007--2016 period\footnote{\url{https://github.com/paigecm/2016-campaign}}. 
We had 300 topics, and we used the distribution over the topics as a representation for the target sentence. We further modeled the context using cosines with such representations for the previous, the current, and the next segment.

\subsubsection{Embeddings (\emph{303 features})}
We also modeled semantics using word embeddings. 
We used the pre-trained 300-dimensional Google News word embeddings by~\citet{DBLP:journals/corr/MikolovLS13} to compute an average embedding vector for the target sentence, and we used the 300 dimensions of that vector. We also modeled the context as the cosine between that vector and the vectors for three segments: the previous, the current, and the following one.

\subsubsection{Contradictions} (\emph{5 features})
Many claims selected for fact-checking contain contradictions to what has been said earlier, as in the example below (from the 3rd presidential debate):\medskip

{\begin{tabular}{rp{116mm}}
Clinton:	& [\ldots] about a potential nuclear competition in Asia, you said, you know, go ahead, enjoy yourselves, folks.	\\
Trump:		& \textbf{I didn't say} nuclear. 
\\\smallskip
\end{tabular}}

\noindent We model this by counting the negations in the target sentence as found in a dictionary of negation cues such as \textit{not}, \textit{didn't}, and \textit{never}. 
We further model the context as the number of such cues in the two neighboring sentences from the same segment and the two neighboring segments.

\subsubsection{Similarity of the sentence to known positive/negative examples (\emph{3 features})}
We used three more features that measure the similarity of the target sentence to other known examples.
The first one computes the maximum over the training sentences of the number of matching words between the target and the training sentence, which is further multiplied by -1 if the latter was not check-worthy. We also used another version of the feature, where we multiplied it by 0 if the speakers were different. A third version took as a training set all claims checked by \emph{PolitiFact}\footnote{\url{http://www.politifact.com/}} (excluding the target sentence).

\subsubsection{Discourse (\emph{20 features})} 
\label{subsubsec:discourse}
We saw above that contradiction can signal the presence of check-worthy claims and contradiction can be expressed by a discourse relation such as \textsc{Contrast}. As other discourse relations such as \textsc{Background}, \textsc{Cause}, and \textsc{Elaboration} can also be useful, we used a discourse parser~\cite{jotycodra} to parse the entire segment. This parser follows the Rhetorical Structure Theory (RST). It produces a hierarchical representation of the discourse by linking first the elementary discourse units with binary discourse relations (indicating also which unit is the \emph{nucleus} and which is the \emph{satellite}), and building up the tree by connecting with the same type of discourse relations the more general cross-sentence nodes until a root node covers all the text. From this tree, we focused on the direct relationship between the target sentence and the other sentences in its segment; this gave rise to 18 contextual indicator features. We further analyzed the internal structure of the target sentence ---how many nuclei and how many satellites it contains---, which gave rise to two sentence-level features.

\subsection{Other Features}


\subsubsection{ClaimBuster-based (\emph{1,045 core features})}
In order to be able to compare our model and features directly to the previous state of the art, we re-implemented, to the best of our ability, the sentence-level features of \emph{ClaimBuster}~\cite{Hassan:15}, namely
TF-IDF-weighted bag of words (998 features), part-of-speech tags (25 features), named entities as recognized by \emph{Alchemy API}\footnote{\url{http://www.ibm.com/watson/alchemy-api.html}}
(20 features), sentiment score from \emph{Alchemy API} (1~feature), and number of tokens in the target sentence (1 feature). 
Apart from providing means of comparison to the state of the art, these features also make a solid contribution to our final system for check-worthiness estimation. 
However, note that we did not have access to the training data of \emph{ClaimBuster}, which is not publicly available, and we thus train on our own dataset. 

\subsubsection{Sentiment (\emph{2 features})}
Some sentences are highly negative, which can signal the presence of an interesting claim to check, as the two following example sentences show (from the 1st and the 2nd presidential debates):
\medskip

{\begin{tabular}{rp{55mm}}
Trump:	& Murders are up.  \\
Clinton:	&  Bullying is up. \\\smallskip
\end{tabular}}

\noindent We used the NRC sentiment lexicon~\cite{Mohammad13} as a source of words and $n$-grams with positive/negative sentiment, and we counted the number of positive and of negative words in the target sentence. These features are different from those in \emph{ClaimBuster}, where these lexicons were not used.

\subsubsection{Named entities (NE) (\emph{1 feature})}
Sentences that contain named entity mentions are more likely to contain a claim that is worth fact-checking as they discuss particular people, organizations, and locations.
Thus, we have a feature that counts the number of named entities in the target sentence; we use the \emph{NLTK toolkit} for named entity recognition~\cite{Loper02nltk:the}.
Unlike the \emph{ClaimBuster} features above, here we only have one feature; we also use a different toolkit for named entity recognition.

\begin{table}[t]
\small
\centering

\begin{tabular}{c@{\hspace{5mm}}c}
\begin{tabular}{lc}
\hline
\textbf{Bias Type} & \textbf{Sample Cues} \\
\hline
Factives & realize, know, discover, learn\\ 
Implicatives & cause, manage, hesitate, neglect \\ 
Assertives & think, believe, imagine, guarantee\\ 
Hedges & approximately, estimate, essentially\\ 
Report-verbs & argue, admit, confirm, express\\ 
Wiki-bias & capture, create, demand, follow\\ 
\hline 
\end{tabular}
&
\begin{tabular}{lc}
\hline
\textbf{Bias Type} & \textbf{Sample Cues} \\
\hline
Modals & can, must, will, shall\\
Negations & neither, without, against, never, none\\
\hline
Strong-subj & admire, afraid, agreeably, apologist\\
Weak-subj & abandon, adaptive, champ, consume\\ 
Positives & accurate, achievements, affirm\\ 
Negatives & abnormal, bankrupt, cheat, conflicts\\
\hline
\end{tabular}
\end{tabular}
\caption{Some cues for various bias types.}
\label{table:bias_types}
\vspace{-2em}
\end{table}

\subsubsection{Linguistic features (13 features)}
We use as features the presence and the frequency of occurrence of linguistic markers such as \textit{factives} and \textit{assertives} from \cite{hooper1974assertive}, \textit{implicatives} from \cite{karttunen1971Implicatives},  \textit{hedges} from \cite{hyland2005metadiscourse}, \textit{Wiki-bias} terms from \cite{Recasens:ACL:13}, \textit{subjectivity} cues from \cite{Riloff:2003:LEP:1119355.1119369}, and 
\textit{sentiment} cues from~\cite{Liu:2005:OOA:1060745.1060797}.\footnote{Most of these bias cues can be found at \url{http://people.mpi-sws.org/~cristian/Biased_language.html}} 
We compute a feature vector 
according to Equation~\eqref{Ling-equation} where for each bias type $B_i$ and answer $A_j$, the frequency of the cues for $B_i$ in $A_j$ is computed and then normalized by the total number of words in $A_j$:
\begin{equation}\label{Ling-equation}
B_i(A_j) = \dfrac{\sum\limits_{cue \in B_i} {count(cue, A_j)}}{\sum\limits_{w_k \in A_j} {count(w_k, A_j)}}
\end{equation}

\newpage
Below we describe these cues in more detail.

\begin{itemize}
\item \emph{Factives} (\emph{1 feature}) \cite{hooper1974assertive}
are verbs that imply the veracity of their complement clause. In \textit{E1}, \textit{know} suggests that ``they will open a second school \dots'' and ``they provide a qualified french education \dots'' are factually true statements.

\begin{description}
\item[\it E1:] 
\begin{itemize}
\textit{\textbf{know}} that they \textit{\textbf{will}} open a second school; and they are a \textit{\textbf{nice}} french school\ldots  I \textit{\textbf{know}} that they \textit{\textbf{provide}} a \textit{\textbf{qualified}} french education and add with that the history and arabic language to be adapted to the qatar. I \textit{\textbf{think}} that's an \textit{\textbf{interesting}} addition.
\end{itemize}
\end{description}

\item \emph{Assertives} (\emph{1 feature}) \cite{hooper1974assertive}
are verbs that imply the veracity of their complement clause with a level of certainty. E.g., in \textit{E1}, \textit{think} indicates some uncertainty, while verbs like \textit{claim} cast doubt on the certainty of their complement clause.

\item \emph{Implicatives} (\emph{1 feature}) \cite{karttunen1971Implicatives}
are verbs that imply the (un)truthfulness of their complement clause, e.g., \textit{decline} and \textit{succeed}.

\item \emph{Hedges} (\emph{1 feature}) \cite{hyland2005metadiscourse}
reduce the person's commitment to the truth, e.g., \textit{may} and \textit{possibly}.

\item \emph{Reporting verbs} (\emph{1 feature})
are used to report a statement from a source, e.g.,~\textit{argue} and \textit{express}.

\item \emph{Wiki-bias cues} (\emph{1 feature}) \cite{Recasens:ACL:13}
are extracted from the NPOV corpus from Wikipedia and cover bias cues (e.g., \textit{provide} in \textit{E1}), and controversial words, such as \textit{abortion} and \textit{execute}. These words are not available in neither of the other bias lexicons. 

\item \emph{Modals} (\emph{1 feature})
are used to change the certainty of the statement (e.g.,~\textit{will} or \textit{can}), make an offer (e.g.,~\textit{shall}), ask permission (e.g.,~\textit{may}), or express an obligation or necessity (e.g., \textit{must}).

\item \emph{Negations} (\emph{1 feature})
are used to deny or make negative statements such as \textit{no}, \textit{never}.

\item \emph{Subjectivity cues} (\emph{2 features}) \cite{Riloff:2003:LEP:1119355.1119369}
are used when expressing personal opinions and feelings. There are \textit{strong} and \textit{weak} cues, e.g., in \textit{E1}, \textit{nice} and \textit{interesting} are \textit{strong}, while \textit{qualified} is \textit{weak}.

\item \emph{Sentiment cues} (\emph{2 features}).
We use \textit{positive} and \textit{negatives}
sentiment cues \cite{Liu:2005:OOA:1060745.1060797}
to model the attitude, thought, and emotions of the speaker. In \textit{E1}, \textit{nice}, \textit{interesting} and \textit{qualified} are positive cues.
\end{itemize}

The above bias and subjectivity cues are mostly single words. Sometimes a multi-word cue (e.g., ``we can guarantee'') can be a stronger signal for user's certainty/uncertainty in their answers. 
We thus further generate multi-word cues \emph{(1 feature)} by combining \emph{implicative}, \emph{assertive}, \emph{factive} and \emph{report} verbs with first person pronouns (\emph{I/we}), \emph{modals} and strong subjective \emph{adverbs}, e.g.,  \emph{I/we+verb} (e.g. ``I believe''), \emph{I/we+adverb+verb} (e.g., ``I certainly know''), \emph{I/we+modal+verb} (e.g., ``we could figure out'') and \emph{I/we+modal+adverb+verb} (e.g., ``we can obviously see'').

\subsubsection{Tense (\emph{1 feature})}
Most of the check-worthy claims mention past events. In order to detect when the speaker is making a reference to the past or is talking about his/her future vision and plans, we include a feature with three values ---indicating whether the text is in past, present or future tense. The feature is extracted in a simplified fashion from the verbal expressions, using POS tags and a list of auxiliary phrases. In particular, we consider a sentence to be in the past tense if it contains a past verb (\emph{VBD}), and in the future tense if it contains \emph{will} or \emph{have to}; otherwise, we assume it to be in the present tense.

\subsubsection{Length (\emph{1 feature})}
Shorter sentences are generally less likely to contain a check-worthy claim.\footnote{One notable exception are short sentences with negations, e.g.,~\emph{Wrong.}, \emph{Nonsense.}, etc.} 
Thus, we have a feature for the length of the sentence in terms of characters.
Note that this feature was not part of the \emph{ClaimBuster} features, as there length was modeled in terms of tokens, but here we do so using characters.

\subsection{Experiments}

\paragraph{Learning Algorithm} We used a feed-forward neural network (FNN) with two hidden layers (with 200 and 50 neurons, respectively) and a softmax output unit for the binary classification.\footnote{Previous work~\cite{pepaRANLP:17} showed that the neural network performs better on the task than support vector machine classifiers.} We used ReLU~\cite{pmlr-v15-glorot11a} as the activation function and we trained the network with Stochastic Gradient Descent~\cite{lecun1998gradient} for 300 epochs with a batch size of 550. We set the L2 regularization to 0.0001, and we kept a constant learning rate of 0.04. We further enhanced the learning process by using a Nesterov's momentum~\cite{sutskever2013importance} of 0.9.

\paragraph{Setting} We trained the models to classify sentences as positive if \emph{one or more media} had fact-checked a claim inside the target sentence, and negative otherwise. We then used the classifier scores to rank the sentences with respect to \emph{check-worthiness}.\footnote{We also tried using ordinal regression, and SVM-perf (an instantiation of SVM-struct), to directly optimize precision, but they performed worse.}
We tuned the parameters and we evaluated the performance using 4-fold cross-validation, using each of the four debates in turn for testing while training on the remaining three.

\paragraph{Implementation Details.}
We used \texttt{gensim}~\cite{rehurek_lrec} for LDA and word embeddings, \texttt{NLTK}~\cite{Loper02nltk:the} for NER and POS tagging, and \texttt{scikit-learn}~\cite{sklearn_api} for deep learning.

\paragraph{Evaluation} We use ranking measures such as \emph{Precision at $k$} ($P@k$) and \emph{Mean Average Precision} (MAP). As Table~\ref{table:per:medium} shows, most media rarely check more than 50 claims per debate,
which means that there is no need to fact-check more than 50 sentences. Thus, we report $P@k$ for $k \in \{5, 10, 20, 50\}$.\footnote{Note that as far as the difference between the P@k metrics (especially between 5 and 10) is in terms of a few sentences, the deviation between them can seem large, while caused by a few correctly/wrongly classified sentences.} 
MAP is the mean of the Average Precision across the four debates. 
%
Finally, we also measure the recall at the $R$-th position of returned sentences for each debate, where $R$ is the number of relevant documents for that debate and the metric is known as $R$-Precision ($R$-Pr). As with MAP, we provide the average across the 4 debates. 

\begin{table}[t]
\small
\centering
\begin{tabular}{l@{ }rrrrrr}
\hline
  \bf Our System 		& \bf MAP & \bf R-Pr & \bf P@5 & \bf P@10 & \bf P@20 & \bf P@50 \\ \hline
  All features 			& \bf0.427 & \bf0.432& \bf0.800 & \bf0.725 & \bf0.713 & \bf0.600\\
  All $\setminus$ discourse	& 0.412 & 0.431 & 0.800 & 0.700 & 0.685 & 0.550\\
  All $\setminus$ context	& 0.385 & 0.390 & 0.550 & 0.500 & 0.550 & 0.540\\
  Only context+discourse	& 0.317 & 0.404 & 0.725 & 0.563 & 0.465 & 0.465\\
\hline
  \bf Reference systems \\
\hline  
  \it Random      		& 0.164 & 0.007 & 0.200 & 0.125 & 0.138 & 0.160\\ 
  \it TF-IDF	  		& 0.314 & 0.333 & 0.550 & 0.475 & 0.413 & 0.360\\
  \it Claimbuster--Platform 	& 0.317 & 0.349 & 0.500 & 0.550 & 0.488 & 0.405\\ 
  \it Claimbuster--Features 	& 0.357 & 0.379 & 0.500 & 0.550 & 0.550 & 0.510\\   
\hline
  \end{tabular}
\caption{\label{table:context}Overall results for check-worthy claims identification, focusing on the impact of the contextual and discourse features.}
\vspace{-1.5em}
\end{table}

\paragraph{Results} Table~\ref{table:context} shows all the results of our claim ranking system with several feature variants. In order to put the numbers in perspective, we also show the results for four increasingly competitive baselines (`Reference Systems'). The first one is a random baseline. It is then followed by an SVM classifier based on a bag-of-words representation with TF-IDF weights estimated on the training data. Then come two versions of the \emph{ClaimBuster} system:
Claimbuster--Platform refers to the performance of \emph{ClaimBuster} using the scores obtained from their online demo,\footnote{
\url{http://idir-server2.uta.edu/claimbuster/demo}} which we accessed on December 20, 2016, and Claimbuster--Features is our re-implementations of \emph{ClaimBuster} using our FNN classifiers trained on our dataset with their features.

\noindent We can see that our system with all features outperforms all reference systems by a large margin for all metrics. The two versions of ClaimBuster also outperform the TF-IDF baseline on most measures. Moreover, our re-implementation of \emph{ClaimBuster} is better than the online platform, especially in terms of MAP\@. This is expected as their system is trained on a different dataset and it may suffer from testing on slightly out-of-domain data.
Our advantage with respect to ClaimBuster implies that the extra information coded in our model, mainly more contextual, structural, and linguistic features, has an important contribution to the final performance. 
 
Rows 2--4 in Table~\ref{table:context} show the effect of the discourse and of the contextual features implemented in our system. The contextual features have a major impact on performance: excluding them yields major drop for all measures, e.g., MAP drops from 0.427 to 0.385, and P@5 drops from 0.800 to 0.550. The discourse features also have an impact, although it is smaller. The most noticeable difference is in the quality at the lower positions in the rank, e.g., P@5 does not vary when removing discourse features, but P@10, P@20 and P@50 all drop by 2.5 to 5 percent points. Finally, row 4 in the table shows that contextual+discourse features alone already yield a competitive system, performing about the same as \emph{Claimbuster--Platform} (which uses no contextual features at all). In Section~\ref{sec:discussion}, we will present a further qualitative description of the results including some examples.

\subsection{Multi-task Learning Experiments}

Unlike the above single-source approaches, in this subsection, we explore a multi-source neural network framework,
in which we try to predict the selections of each and every fact-checking organization simultaneously. We show that, even when the goal is to mimic the selection strategy of one particular fact-checking organization, it is beneficial to leverage on the selection choices by multiple such organizations.

\paragraph{Setting} We approach the task of check-worthiness prediction using the same features, while at the same time modeling the problem as multi-task learning, using different sources of annotation over the same training dataset. As a result, we can learn to mimic the selection strategy of each and every of these individual sources.  As we have explained above, in our dataset the individual judgments come from nine independent fact-checking organizations, and we thus predict the selection choices of each of them in isolation plus a collective label \emph{ANY}, which indicates whether at least one source would judge that claim as check-worthy. 

\begin{figure}[t]
\includegraphics[width=350pt]{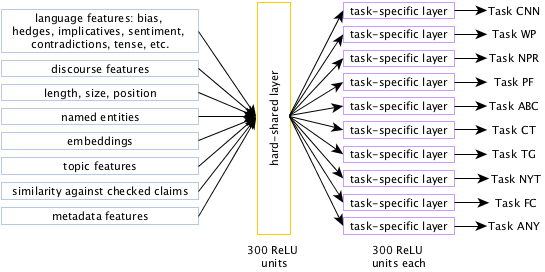}
\caption{The architecture of the full neural multi-source learning model, predicting the selection choices of each of the nine individual sources (tasks) and of one cumulative source: \emph{task ANY}.}
\label{figure:architecture}
\end{figure}

\paragraph{Architecture}
Figure~\ref{figure:architecture} illustrates the architecture of the full neural multi-source learning model, which predicts the selection choices of each of the nine individual sources (tasks) and of the special cumulative source: \emph{task ANY}. 
There is a hidden layer (of size 300) that is shared between all ten tasks. Then, each task has its own task-specific hidden layer (each of size 300). Finally, each task-specific layer is followed by an output layer: a single sigmoid unit that provides the prediction of whether the utterance was fact-checked by the corresponding source. Eventually, we make use of the probability of the prediction to prioritize claims for fact-checking. During training, each task modifies the weights of both its own task-specific layer and of the shared layer.
For our neural network architecture, we used ReLU units, Stochastic Gradient Descent with Nesterov momentum of 0.7, iterating for 100 epochs with batches of size 500 and a learning rate of 0.08.

This kind of neural network architecture for multi-task learning is known in the literature as \emph{hard parameter sharing}~\cite{Caruana93multitasklearning}, and it can greatly reduce the risk of overfitting. In particular, it has been shown that the risk of overfitting the shared parameters in the hidden layer is an order $n$ smaller than overfitting the task-specific parameters in the output layers, where $n$ is the number of tasks at hand~\cite{Baxter1997}.
The input to our neural network consists of the various domain-specific features that have been previously described.

\paragraph{Implementation Details} 
We implemented the neural network using Keras.
We tried adding more shared and task-specific layers as well as having some task-specific layers linked directly to the input, but we eventually settled on the architecture in  Figure~\ref{figure:architecture}. We also tried to optimize directly for average precision and adding loss weights to \emph{task ANY}, but using the standard binary cross-entropy loss yielded the best results.

\renewcommand{\tabcolsep}{2pt}
\begin{table}
\small
\centering
\begin{tabular}{c@{\hspace{5mm}}c}

\begin{tabular}{lrrrrrr}
\hline
    \bf Model & \bf MAP & \bf R-Pr & \bf P@5 & \bf P@10 & \bf P@20 & \bf P@50 \\
    \hline
\\
\multicolumn{7}{l}{\bf ABC} \\
\hline
      \mA & 0.097 & 0.112 & 0.250 & 0.175 & 0.162 & 0.100 \\
  \bf \mD & \bf 0.119 & \bf 0.157 & \bf 0.333 & \bf 0.225 & \bf 0.217 & \bf 0.122 \\
  \bf \mE & \bf 0.118 & \bf 0.160 & \bf 0.300 & \bf 0.233 & \bf 0.229 & \bf 0.132 \\
\hline
\\
\multicolumn{7}{l}{\bf The Washington Post (WP)} \\
\hline
      \mA & 0.106 & 0.110 & 0.150 & 0.100 & 0.112 & 0.110 \\
  \bf \mD & \bf 0.127 & \bf 0.127 & \bf 0.350 & \bf 0.233 & \bf 0.162 & \bf 0.123 \\
  \bf \mE & \bf 0.130 & \bf 0.129 & \bf 0.350 & \bf 0.250 & \bf 0.171 & \bf \emph{0.110} \\
\hline
\\
\multicolumn{7}{l}{\bf CNN} \\
\hline
      \mA & 0.087 & 0.091 & 0.250 & 0.150 & 0.121 & 0.090 \\
  \bf \mD & \bf 0.113 & \bf 0.132 & \bf \emph{0.250} & \bf 0.208 & \bf 0.183 & \bf 0.140 \\
  \bf \mE & \bf 0.109 & \bf 0.126 & 0.167 & \bf 0.200 & \bf 0.167 & \bf 0.128 \\
\hline
\\
\multicolumn{7}{l}{\bf FactCheck (FC)} \\
\hline
      \mA & 0.084 & 0.114 & 0.117 & 0.125 & 0.088 & 0.100 \\
  \bf \mD & \bf 0.105 & \bf 0.136 & \bf 0.250 & \bf 0.175 & \bf 0.146 & \bf 0.118 \\
  \bf \mE & \bf 0.117 & 0.110 & \bf 0.333 & \bf 0.242 & \bf 0.196 & \bf 0.107 \\
\hline
\\
\multicolumn{7}{l}{\bf PolitiFact} \\
\hline
           \mA & 0.201 & 0.278 & 0.250 & 0.250 & 0.262 & 0.262 \\
  \textbf{\mD} & \bf 0.209 & 0.258 & \bf 0.400 & \bf 0.367 & \bf 0.317 & \bf 0.270 \\
  \textbf{\mE} & \bf 0.210 & 0.252 & \bf 0.500 & \bf 0.350 & \bf 0.333 & \bf 0.272 \\
\hline
\end{tabular}

& 

\begin{tabular}{lrrrrrr}
\hline
\bf Model & \bf MAP & \bf R-Pr & \bf P@5 & \bf P@10 & \bf P@20 & \bf P@50 \\
    \hline
\\
\multicolumn{7}{l}{\bf NPR} \\
\hline
      \mA & 0.175 & 0.195 & 0.250 & 0.250 & 0.283 & 0.228 \\
  \bf \mD & \bf 0.186 & \bf 0.210 & \bf 0.333 & \bf 0.342 & \bf 0.300 & \bf 0.245 \\
  \bf \mE & \bf 0.180 & \bf 0.207 & \bf 0.333 & \bf 0.283 & 0.250 & 0.227 \\
\hline
\\
\multicolumn{7}{l}{\bf The Guardian (TG)} \\
\hline
      \mA & 0.127 & 0.174 & 0.200 & 0.150 & 0.196 & 0.178 \\
  \bf \mD & \bf 0.133 & \bf 0.199 & 0.183 & \bf 0.175 & \bf 0.192 & \bf 0.193 \\
  \bf \mE & \bf 0.130 & 0.159 & \bf \bf 0.217 & \bf 0.175 & \bf 0.200 & 0.167 \\
\hline
\\
\multicolumn{7}{l}{\bf Chicago Tribune (CT)} \\
\hline
      \mA & 0.079 & 0.110 & 0.100 & 0.100 & 0.125 & 0.075 \\
  \bf \mD & \bf 0.081 & 0.090 & \bf \emph{0.100} & \bf 0.133 & 0.104 & \bf 0.082 \\
  \bf \mE & \bf 0.087 & 0.087 & \bf 0.133 & \bf \emph{0.100} & 0.108 & \bf 0.093 \\
\hline
\\
\multicolumn{7}{l}{\bf The New York Times (NYT)} \\
\hline
      \mA & 0.187 & 0.221 & 0.350 & 0.325 & 0.238 & 0.192 \\
  \bf \mD & 0.150 & 0.213 & 0.233 & 0.200 & 0.196 & 0.180 \\
  \bf \mE & 0.147 & 0.197 & 0.200 & 0.167 & 0.158 & 0.162 \\
\hline
\\\\\\\\\\
  \end{tabular}
    \end{tabular}
\caption{Evaluation results for each of the nine fact-checking organizations as a target to mimic. Shown are results for single-source baselines and for multi-task learning. The improvements over the corresponding baselines are marked in bold.}
\label{table:multitask:results}
 \end{table}

\begin{table}[h!]
\small
\centering
\begin{tabular}{lrrrrrr}
\hline
\bf Model & \bf MAP & \bf R-Pr & \bf P@5 & \bf P@10 & \bf P@20 & \bf P@50 \\
\hline
\mCBP 		& 0.090 & 0.138 & 0.144 & 0.143 & 0.121 & 0.117\\
\mRANLPsingle	& 0.120 & 0.142 & 0.228 & 0.206 & 0.179 & 0.137\\
\mRANLP		& 0.128 & 0.225 & 0.194 & 0.186 & 0.178 & 0.153\\
\hline
\mA \emph{(embed.)} & 0.058 & 0.065 & 0.055 & 0.055 & 0.068 & 0.072\\
\mCBfeats 	  & 0.072 & 0.077 & 0.106 & 0.076 & 0.081 & 0.079\\
\hline
      \mA & 0.127 & 0.156 & 0.213 & 0.181 & 0.176 & 0.148 \\
  \bf \mD & \bf 0.136 & \bf 0.169 & \bf 0.270 & \bf 0.229 & \bf 0.202 & \bf 0.164 \\
  \bf \mE & \bf 0.136 & \bf 0.159 & \bf 0.281 & \bf 0.222 & \bf 0.201 & \bf 0.155 \\
\hline
      \mB & 0.125 & 0.153 & 0.204 & 0.197 & 0.175 & 0.153 \\    
      \mC & \bf 0.130 & 0.153 & \bf 0.237 & \bf 0.220 & \bf 0.184 & \bf \emph{0.148} \\
\hline
\end{tabular}
\caption{Evaluation results averaged over nine fact-checking organizations. The improvements over \emph{singleton} are in bold.}
\vspace{-1em}

\label{tab:avg-map}
\end{table}  

\paragraph{Results} As before, we perform 4-fold cross-validation, where each time we leave one debate out for testing. Moreover, in order to stabilize the results, we repeat each experiment three times with different random seeds, and we report the average over these three reruns.\footnote{Having multiple reruns is a standard procedure to stabilize an optimization algorithm that is sensitive to the random seed, e.g., this strategy has been argued for when using MERT for tuning hyper-parameters in Statistical Machine Translation \cite{Foster:2009:SME}.} We should note that in most cases this was not really needed, as the standard deviation for the reruns was generally tiny: 0.001 or less, absolute. 

Table~\ref{table:multitask:results} presents the results, with all evaluation metrics, when predicting each of the nine sources.
We experiment with three different configurations of the model described in the previous section. 
All of them aim at learning to mimic the selection choices by one single fact-checking organization (source). The first one is a single-task baseline \mA where a separate neural network is trained for each source. 
The other two are multi-task learning configurations:
\mD trains to predict labels for each of the nine tasks, one for each fact-checker;
and \mE trains to predict labels for each of the nine tasks (one for each fact-checker), and also for \emph{task ANY} (as shown in Figure~\ref{figure:architecture}).
We can see in Table~\ref{table:multitask:results} that, for most of the sources, multi-task learning improves over the single-source system. 
The results of the multi-task variations that improve over the single baseline are boldfaced in the table.
The improvements are consistent across evaluation metrics and vary largely depending on the source and the metric.
One notable exception is NYT, for which the single-task learning shows the highest scores. We hypothesize that the network has found some distinctive features of NYT, which make it easy to predict. These relations are blurred when we try to optimize for multiple tasks at once. However, it is important to state that removing NYT from the learning targets worsens the results for the other sources, i.e., it carries some important relations that are worth modeling.

The first three rows of Table~\ref{tab:avg-map} present the same results but averaged over the nine sources. Again, we can see that  multi-task learning yields sizable improvement over the single-task learning baseline for all evaluation measures. 
Another conclusion that can be drawn is that including the task \emph{any} 
does not help to improve the multi-task model.  This is probably due to the fact that this information is already contained in the multi-task model with nine distinct sources only.
The last two rows in Table~\ref{tab:avg-map} present two additional variants of the model: the single-task learning \mB system, which is trained on the union of the selected sentences by all nine fact-checkers to predict the target fact-checker only; and 
the system \mC that predicts labels for two tasks: (\emph{i})~for the target fact-checker, and (\emph{ii})~for \emph{task ANY}. 

We can see that the model \mB performs comparably to the \mA baseline, thus being clearly inferior than the multi-task learning variants. 
Finally, \mC is also better than the single-task learning variants, but it falls short compared to the other multi-task learning variants. Including output units for all nine individual media seems crucial for getting advantage of the multi-task learning, i.e., considering only an extra output prediction node for \emph{task ANY} is not enough.


\section{Fact-Checking}
\label{sec:cqa}



With the ever growing amount of unreliable content online, veracity will almost certainly become an important component of question answering systems in the future.
In this section, we focus on fact-checking in the context of community question answering (cQA), i.e.,~predicting whether an answer to a given question is likely to be true. 
This aspect has been ignored, e.g.,~in recent cQA tasks at NTCIR and SemEval~\cite{Ishikawa:10,nakov-EtAl:2015:SemEval,nakov-EtAl:2016:SemEval,SemEval-2017:task3}, where an answer is considered as \good\ if it tries to address the question, irrespective of its veracity.
Yet, veracity is an important aspect, as high-quality automatic fact-checking can offer a better experience to the users of cQA systems; e.g., a possible application scenario would be that in which the user could be presented with a ranking of all good answers accompanied by veracity scores, where low scores would warn her not to completely trust the answer or to double-check it. 

\begin{figure}
\begin{tabular}{rp{125mm}}
$q$:	& 
If wife is under her husband's sponsorship and is willing to come Qatar on visit, how long she can stay after extending the visa every month? I have heard it's not possible to extend visit visa more than 6 months? \ldots
\\\\
$a_1$:	& Maximum period is 9 Months....	\\
$a_2$:	& 6 months maximum	\\
$a_3$:	& This has been answered in QL so many times. Please do search for information regarding this. BTW answer is 6 months.	
\end{tabular}
\caption{\label{fig:example}Example from the Qatar Living forum.}
\label{fig:example1}
\end{figure}

Figure~\ref{fig:example1} presents an excerpt of an example from the Qatar Living forum, with one question ($q$) and three plausible answers ($a_1-a_3$) selected from a longer thread. According to the SemEval-2016 Task 3 annotation instructions \cite{nakov-EtAl:2016:SemEval}, all three answers are considered \good\ since they address the question. Nevertheless, $a_1$ contains false information, while $a_2$ and $a_3$ are true,\footnote{One could also guess that answers $a_2$ and $a_3$ are more likely to be true from the fact that the \emph{6 months} answer fragment appears many times in the current thread (it also happens to appear more often in related threads as well). While these observations serve as the basis for useful features for classification, the real verification for a gold standard annotation requires finding support from a reliable external information source: in this case, an official government information portal.} as can be checked on an official governmental website.\footnote{\url{https://www.moi.gov.qa/site/english/departments/PassportDept/news/2011/01/03/23385.html}}


\subsection{Data}

\label{sec:corpus}
We use the CQA-QL-FACT dataset, which stresses the difference between (a)~distinguishing a good vs.\ a bad answer, and (b)~distinguishing between a factually true vs.\ a factually false one. 
We added the factuality annotations on top of the CQA-QL-2016 dataset from the SemEval-2016 Task 3 on community Question Answering~\cite{nakov-EtAl:2016:SemEval}.
%
In CQA-QL-2016, the data is organized in question--answer threads extracted from the Qatar Living forum. Each question has a subject, a body, and metadata: ID, category (e.g., \emph{Computers and Internet}, \emph{Education}, and \emph{Moving to Qatar}), date and time of posting, and user name. 

First, we annotated the questions using the following labels:
\begin{itemize}
\item \textsc{Factual}:
The question is asking for factual information, which can be answered by checking various information sources, and it is not ambiguous.
\item \textsc{Opinion}:
The question asks for an opinion or an advice, not for a fact.
\item \textsc{Socializing}: 
Not a real question, but rather socializing/chatting.
This can also mean expressing an opinion or sharing some information without really asking anything of general interest.
\end{itemize}
We annotated 1,982 questions, with the above factuality labels.
We ended up with 625 instances that contain multiple questions, which we excluded from further analysis. Table~\ref{table:question-labels-distribution} shows the annotation results for the remaining 1,357 questions, including examples.

\begin{table}
\small
\centering
\begin{tabular}{l	r @{\hspace{5mm}}p{60mm}}
\toprule
\bf Label & \bf \#Questions & \bf Example\\ \midrule
\textsc{Factual} & 373	& What is Ooredoo customer service number?\\
\textsc{Opinion} & 689	& Can anyone recommend a good Vet in Doha? \\
\textsc{Socializing} & 295	& What was your first car? \\ 
\bottomrule
\end{tabular}
\caption{Distribution of factuality labels in the annotated questions (1,357 in total).}
\label{table:question-labels-distribution}
\vspace{-1.5em}
\end{table}

%
Next, we annotated for veracity the answers to the factual questions.
We only annotated the originally judged as \good\  answers (ignoring both \bad\ and \potential), and we used the following labels:
%

%
%
%

\begin{itemize}
\item \textsc{Factual - True}: 
The answer is True and this can be verified using an external resource.
(\emph{q:~``I wanted to know if there were any specific shots and vaccinations I should get before coming over [to Doha].''; a: ``Yes there are; though it varies depending on which country you come from. In the UK; the doctor has a list of all countries and the vaccinations needed for each.''}).\footnote{This can be verified at \url{https://wwwnc.cdc.gov/travel/destinations/traveler/none/qatar}}

\item \textsc{Factual - False}:
The answer gives a factual response, but it is false.
(\emph{q:~``Can I bring my pitbulls to Qatar?'', a:~``Yes you can bring it but be careful this kind of dog is very dangerous.''}).\footnote{The answer is not true because pitbulls are included in the list of banned breeds in Qatar: \url{http://canvethospital.com/pet-relocation/banned-dog-breed-list-qatar-2015/}}

\item \textsc{Factual - Partially True}:
We could only verify part of the answer.
(\emph{q:~``I will be relocating from the UK to Qatar [\ldots] is there a league or TT clubs / nights in Doha?'', a: ``Visit Qatar Bowling Center during thursday and friday and you'll find people playing TT there.''}).\footnote{The place has table tennis, but we do not know on which days: \url{https://www.qatarbowlingfederation.com/bowling-center/}}

\item \textsc{Factual - Conditionally True}:
The answer is True in some cases, and False in others, depending on some conditions that the answer does not mention.
(\emph{q: ``My wife does not have NOC from Qatar Airways; but we are married now so can i bring her legally on my family visa as her husband?'', a: ``Yes you can.''}).\footnote{This answer can be true, but this depends upon some conditions: \url{http://www.onlineqatar.com/info/dependent-family-visa.aspx}}

\item \textsc{Factual - Responder Unsure}:
The person giving the answer is not sure about the veracity of his/her statement.
(e.g., ``\emph{Possible only if government employed. That's what I heard.}'')

\item \textsc{NonFactual}:  The answer is not factual.
It could be an opinion, an advice, etc.
that cannot be verified.
(e.g., ``\emph{Its better to buy a new one.}'')

\end{itemize}



We further discarded items whose factuality was very time-sensitive (e.g., ``\emph{It is Friday tomorrow.}'', ``\emph{It was raining last week.}'')\footnote{Arguably, many answers are somewhat time-sensitive, e.g., ``\emph{There is an IKEA in Doha.}'' is true only after IKEA opened, but not before that. In such cases, we just used the present situation as a point of reference.},
or for which the annotators were unsure.

\begin{table}
\small
\centering
\begin{tabular}{lc@{\hspace{3mm}}	lr}
\hline
\textbf{Coarse-Grained Label} 	& \textbf{Answers} & {\bf Fine-Grained Label} &   \textbf{Answers}\\
\hline
$+$ \bf \textsc{Positive}	& \bf 128	& $+$ \textsc{Factual - True} 		& 128	\\\hline
$-$ \bf \textsc{Negative}	& \bf 121	& $-$ \textsc{Factual - False} 		& 22	\\
				&		& $-$ \textsc{Factual - Partially True} & 38 \\
				&		& $-$ \textsc{Factual - Conditionally True} & 16 \\
				&		& $-$ \textsc{Factual - Responder Unsure} & 26 \\
				&		& $-$ \textsc{NonFactual}		& 19 \\
\hline
\end{tabular}
%
%
\caption{Distribution of the positive and the negative answers (i.e., the two classes we predict) and of the fine-grained labels.}
\label{table:comment-labels-distribution}
\vspace{-1em}
\end{table}

\noindent We considered all questions from the \textsc{Dev} and the \textsc{Test} partitions of the CQA-QL-2016 dataset.
We targeted very high quality, and thus we did not crowdsource the annotation, as pilot annotations showed that the task was very difficult and that it was not possible to guarantee that Turkers would do all the necessary verification, e.g.,~gather evidence from trusted sources. Instead, all examples were first annotated independently by four annotators,
%
and then they discussed \emph{each example} in detail to come up with a final label.
We ended up with 249 \textsc{Good} answers\footnote{This is comparable in size to other fact-checking datasets, e.g., \citet{Ma:2015:DRU} used 226 rumors, and \citet{popat2016credibility} had 100 Wiki hoaxes.} to 71 different questions, which we annotated for factuality: 128 \textsc{Positive} and 121 \textsc{Negative} examples. See  Table~\ref{table:comment-labels-distribution} for details.

\subsection{Modeling Context and Discourse}

We model the \emph{context} of an answer with respect to the entire answer thread in which it occurs, and with respect to other high-quality posts from the entire Qatar Living forum. We further use \emph{discourse} features as in Section~\ref{subsubsec:discourse}.


\subsubsection{\forumRankSameUser\ (\emph{5 features})}
We use the cosine similarity between an answer- and a thread-vector of all \good \ answers using Qatar Living embeddings. For this purpose, we use 100-dimensional in-domain word embeddings~\cite{SemEval2016:task3:SemanticZ}, which were trained using \textsc{word2vec}~\cite{mikolov-yih-zweig:2013:NAACL-HLT} on a large dump of Qatar Living data (2M answers).\footnote{Available at \url{http://alt.qcri.org/semeval2016/task3/data/uploads/QL-unannotated-data-subtaskA.xml.zip}}
The idea is that if an answer is similar to other answers in the thread, it is more likely to be true. 
To this, we add 
thread-level features related to the rank of the answer in the thread:
	\Ni the reciprocal rank of the answer in the thread and
	\Nii percentile of answer's rank in the thread. 
    As there are exactly ten answers per thread in the dataset, the first answer gets the score of 1.0, the second one gets 0.9, the next one gets 0.8, and so on. 
We calculate these two ranking features twice: once for the full list of answers, and once for the list of good answers only.







\subsubsection{\forumIRQLsplit\ (\emph{60 features})}
We further collect supporting evidence from all threads in the Qatar Living forum. To do this, we query a search engine, limiting the search to the forum only. See Section~\ref{sec:web:features} for more detail about how the search for evidence on the Web is performed and what features are calculated.


\begin{table}
\small
\centering
\begin{tabular}{l@{\hspace{1mm}}p{120mm}}
\hline
\textbf{Question}:	& does anyone know if there is a french speaking nursery in doha?\\
\textbf{Answer}:	& there is a french school here. don't know the ages but my neighbor's 3 yr old goes there\ldots \\
\hline
\end{tabular}

\begin{tabular}{lcccp{105mm}}
\multicolumn{5}{l}{\textbf{Best Matched Sentence for Q\&A:} there is a french school here.}\\
\hline
\\
\hline
\textbf{Post Id} & \textbf{sNo} & \textbf{R1} & \textbf{R2} & \textbf{Sentence}\\
\hline
35639076 & 15 & 1 & 10 & the pre-school follows the english program but also gives french and arabic lessons.\\ 
32448901 & 4 & 2 & 11 & france bought the property in 1952 and since 1981 it has been home to the french institute.\\ 
31704366 & 7 & 3 & 1 & they include one indian school, two french, seven following the british curriculum\ldots \\
27971261 & 6 & 4 & 4 & the new schools include six qatari, four indian, two british, two american and a finnish\ldots \\
\hline
\\
\end{tabular}
\caption{Sample of sentences from high-quality posts automatically extracted to support the answer \textit{A}. \textit{sNo} is the sentence's sequential number in the post, \textit{R1} and \textit{R2} are the ranks of the target sentences based on entailment and similarity, respectively.}
\label{table:article_supports}
\vspace{-1.5em}
\end{table}

\subsubsection{\forumHighQualPosts}
\label{Inter-thread-evidence}

Among the $60k$ active users of the Qatar Living forum, there is a community of 38 trusted users, who have written $5.2k$
high-quality articles on topics that attract a lot of interest, e.g., issues related to visas, work legislation, etc.
%
%
We try to verify the answers against these high-quality posts.
\Ni Since an answer can combine both relevant and irrelevant information with respect to its question, we first generate a query against a search engine for each Q\&A.
\Nii We then compute cosines between the query and the sentences in the high-quality posts, and we select the $k$-best matches. 
\Niii Finally, we compute textual entailment scores \cite{Kouylekov:2010} for the answer given the $k$-best matches, which we then use as features.
An example is shown in Table~\ref{table:article_supports}.

\subsubsection{Discourse features} We use the same discourse features as for the claim identification task (cf. Section~\ref{subsubsec:discourse}).

\subsection{Other Features}


\subsubsection{\contentLingf}\label{LinguisticAnalysis}
Forum users, consciously or not, often put linguistic markers in their answers, which can signal the degree of the user's certainty in the veracity of what they say. We thus use the linguistic features from the previous task (see above).


\subsubsection{\contentCredibility\ (\emph{31 features})}
We use features 
that have been previously proposed for credibility detection \cite{Castillo:2011:ICT:1963405.1963500}:
number of URLs/images/emails/phone numbers;
number of tokens/sentences;
average number of tokens;
number of positive/negative smileys;
number of single/double/triple exclamation/in\-terrogation symbols.
To this set, we further add number of interrogative sentences;
number of nouns/verbs/adjectives/adverbs/pronouns;
and number of words that are not in word2vec's Google News vocabulary (such OOV words could signal slang, foreign language, etc.)
We also use the number of 1st, 2nd, 3rd person pronouns in the comments: \Ni in absolute number, and also \Nii normalized by the total number of pronouns in the comment. The latter is also a feature. 

\begin{table}
\small
\centering
\begin{tabular}{@{}l@{}c@{}c p{105mm} l@{}}
\hline
\multicolumn{4}{l}{\textbf{Question:} Hi; Just wanted to confirm Qatar's National Day. Is it 18th of December? Thanks.}\\
\multicolumn{4}{l}{\textbf{Answer:} yes; it is 18th Dec.}\\
\hline
\multicolumn{4}{l}{\textbf{Query generated from Q\&A:} \texttt{"National Day" "Qatar" National December Day confirm wanted}}\\
\hline
\\
\hline
& \textbf{Qatar-} & \textbf{Source} &\\
\textbf{URL} & \textbf{related?} & \textbf{type} & \textbf{Snippet}\\
\hline
\url{qppstudio.net} & No & Other & Public holidays and national \ldots the world's source of Public holidays information\\ 
\url{dohanews.co} & Yes & Reputed & culture and more in and around Qatar \ldots The documentary features human interest pieces that incorporate the day-to-day lives of Qatar residents\\ 
\url{iloveqatar.net} & Yes  & Forum & Qatar National Day - Short Info \ldots the date of December 18 is celebrated each year as the National Day of Qatar\ldots\\
\url{cnn.com} & No & Reputed & The 2022 World Cup final in Qatar will be held on December 18 \ldots Qatar will be held on December 18 -- the Gulf state's national day.  Confirm. U.S \ldots\\
\url{icassociat} & No & Other & In partnership with ProEvent Qatar, ICA can confirm that the World Stars \\ 
\url{ion.co.uk} & & & will be led on the 17 December, World Stars vs. Qatar Stars - Qatar National Day.\\
\hline
\\
\end{tabular}
\caption{Sample snippets returned by a search engine for a given query generated from a Q\&A pair.}
\label{table:ir_example}
\vspace{-1.5em}
\end{table}

\subsubsection{\extIRFBingWebpageQataronlySplitted\  (\emph{60~features})}
\label{sec:web:features}
We tried to verify whether an answer's claim is factually true by searching for supporting information on the Web. We started with the concatenation of an answer to the question that heads the respective thread. Then, following \cite{potthast2013overview}, we extracted nouns, verbs and adjectives, sorted by TF-IDF (we computed IDF on the Qatar Living dump). We further extracted and added the named entities from the text and we generated a query of 5-10 words. If we did not obtain ten results, we dropped some terms 
and we tried again. 

\noindent We automatically queried Bing and Google, and we extracted features from the resulting pages, considering Qatar-related websites only. An example is shown in Table~\ref{table:ir_example}.
Based on the results, we calculated similarities: \Ni cosine with TF-IDF vectors, \Nii cosine using Qatar Living embeddings, and \Niii containment~\cite{lyon2001detecting}. We calculated these similarities between, on the one side, \Ni the question or \Nii the answer or \Niii the question--answer pair, vs.\ on the other side, \Ni the snippets or \Nii the web pages. To calculate the similarity to a webpage, we first converted the page to a list of rolling sentence triplets, then we calculated the score of the Q/A/Q-A vs.\ this triplet, and finally we took the average and also the maximum similarity over these triplets.
Now, as we had up to ten Web results, we further took the maximum and the average over all the above features over the returned Qatar-related pages.
We created three copies of each feature, depending on whether it came from a \Ni reputed source (e.g., news, government websites, official sites of companies, etc.), from a \Nii forum type site (forums, reviews, social media), or \Niii from some other type of websites.

\subsubsection{Embeddings}	 (\emph{260~features})
Finally, we used as features the embeddings of the claim (i.e., the answer), of the best-scoring snippet, and of the best-scoring sentence triplet from a webpage. We calculated these embeddings using long short-term memory (LSTM) representations, which we trained for the task as part of a deep neural network (NN). We also used a task-specific embedding of the question and of the answer together with all the above evidence about it, which comes from the last hidden layer of the neural network.

\subsection{Classification Model}

\begin{figure}
\centering
\begin{subfigure}{\textwidth}
\centering
\includegraphics[width=0.65\textwidth]{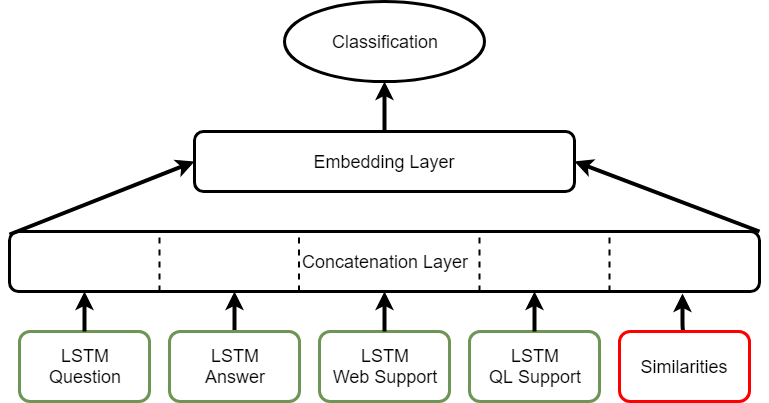}\vspace*{4mm}
\caption{General overview of the architecture.}
\label{fig:NN-top}
\end{subfigure}
\begin{subfigure}{\textwidth}
\centering
\includegraphics[width=0.93\textwidth]{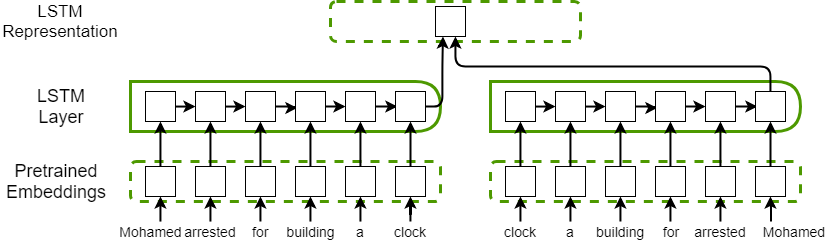}
\caption{Detailed LSTM representation.}
\label{fig:NN-bottom}
\end{subfigure}

\caption{Our NN architecture for fact-checking in cQA.
Each green box in~\ref{fig:NN-top} consists of the bi-LSTM structure in~\ref{fig:NN-bottom}.
\label{fig:NN}}
\vspace{-1.2em}
\end{figure}

Our model combines an LSTM-based neural network with kernel-based support vector machines.
In particular, we use a bi-LSTM recurrent neural network to train abstract feature representations of the examples. We then feed these representations into a kernel-based SVM, together with other features. 
The architecture is shown in Figure~\ref{fig:NN-top}. We have five LSTM 
sub-networks, one for each of the text sources from two search engines: \emph{Claim}, \emph{Google Web page}, \emph{Google snippet}, \emph{Bing Web page}, and \emph{Bing snippet}. We feed the claim (i.e., the answer) into the neural network as-is. As we can have multiple snippets, we only use the best-matching one as described above. Similarly, we only use a single best-matching triple of consecutive sentences from a Web page. We further feed the neural network with the similarity features described above.
All these vectors are concatenated and fully connected to a much more compact hidden layer that captures the task-specific embeddings. This layer is connected to a softmax output unit that classifies the claim as true or false. Figure~\ref{fig:NN-bottom} shows the general architecture of each of the LSTM components.
The input text is transformed into a sequence of word embeddings, which are then passed to the bidirectional LSTM layer to obtain a representation for the input text.

Next, we extract the last three layers of the neural network ---\Ni the concatenation layer, \Nii the embedding layer, and \Niii the classification node--- and we feed them into an SVM with a radial basis function kernel (RBF). In this way, we use the neural network to train task-specific embeddings of the input text fragments, and also of the entire input example. Ultimately, this yields a combination of deep learning and task-specific embeddings with RBF kernels.
\subsection{Experiments} 

\subsubsection{Question Classification}

Table~\ref{table:subtaskA} shows the results of our run for classification of the three question categories (\textsc{Factual}, \textsc{Opinion}, \textsc{Socializing}), using an SVM with bag-of-words and some other features. We can see a 10-point absolute improvement over the baseline, which means the task is feasible. This also leaves plenty of space for further improvement, which is beyond the scope of this work. Instead, below we focus on the more interesting task of checking the factuality of \emph{Good} answers to \emph{Factual} questions.

\begin{table}
\centering
\begin{tabular}{lccccc}
\toprule
\textbf{System} & \textbf{Accuracy} \\ \midrule
 Baseline: All \textsc{Opinion} (majority class) & 50.7 \\ \midrule
 Our pilot: SVM, bag-of-words & 62.0 \\
 Our pilot: SVM, text features & 60.3 \\
\bottomrule
 \\
\end{tabular}
\caption{Baseline vs. our pilot SVM, predicting one of three classes of the questions (factual vs. opinion vs. just socializing).}
\label{table:subtaskA}
\vspace{-2em}
\end{table}

\subsubsection{Answer Classification}
\label{ssub:answer-classif}

\paragraph{Setting and Evaluation} We perform leave-one-thread-out cross validation, where each time we exclude and use for testing one of the 71 questions together with all its answers. This is done in order to respect the structure of the threads when splitting the data. We report Accuracy, Precision, Recall, and F$_1$ for the classification setting. 

We used a bidirectional LSTM with 25 units and a hard-sigmoid activation, which we trained using an RMSprop optimizer with 0.001 initial learning rate, L2 regularization with $\lambda$=0.1, and 0.5 dropout after the LSTM layers. The size of the hidden layer was 60 with \emph{tanh} activations. We used a batch of 32 and we trained for 400 epochs. Similarly to the bi-LSTM layers, we used an $l_2$ regularizer with $\lambda$ = 0.01 and dropout with a probability of 0.3.

%
For the SVM, we used grid search to find the best parameters for the parameters $c$ and $\gamma$. 
We optimized the SVM for classification Accuracy. 

\paragraph{Results} Table~\ref{table:cqa-context} shows the results from our experiments for several feature combinations and for two baselines. 
First, we can see that our system with all features performs better than the baseline for Accuracy. The ablation study shows the importance of the context and of the discourse features. When we exclude the discourse and the contextual features, the accuracy drops from 0.683 to 0.659 and 0.574, respectively. When both the context and the discourse features are excluded, the accuracy drops even further, to 0.542. The F$_1$ 
results are consistent with this trend. This is similar to the trend for check-worthiness estimation (cf. Table~\ref{table:context}). Finally, using the discourse and the contextual features, without any other features, yields an accuracy of 0.635, which is quite competitive. 
Overall, these results show the importance of the contextual and of the discourse features for the fact-checking task, with the former being more important than the latter.

\begin{table}
\small
\centering
\begin{tabular}{l@{ }ccccc}
\hline
  \bf Our System & \bf Accuracy & \bf Precision & \bf Recall & \bf F$_1$ 
  \\ \hline
  All information sources 		& 0.683 & 0.693 & 0.688 & 0.690 
  \\
  All $\setminus$ discourse  		& 0.659 & 0.675 & 0.648 & 0.661 
  \\  
  All $\setminus$ context 		& 0.574 & 0.583 & 0.602 & 0.592 
  \\
  All $\setminus$ discourse and context & 0.542 & 0.554 & 0.563 & 0.558 
  \\
  Only context+discourse 		& 0.635 & 0.621 & 0.742 & 0.676 
  \\
\hline
  \bf Baseline \\
\hline  
  \it All positive (majority class)     & 0.514 & 0.514 & 1.000 & 0.679 
  \\ 
\hline
  \end{tabular}
\caption{Fact-checking the answers in a cQA forum, focusing on the impact of the contextual and discourse features. 
}
\label{table:cqa-context}
\vspace{-1.5em}
\end{table}


\section{Discussion}
\label{sec:discussion}


Here we look at some examples that illustrate how context and discourse help for our two tasks.

\subsection{Impact of Context}

First, we give some examples where the use of contextual information yields the correct prediction for the check-worthiness task (Section~\ref{sec:debates}). In each of these examples, there is a particular contextual feature type that turned out to be critical for making the correct prediction, namely that these are check-worthy sentences (they were all misclassified as not check-worthy when excluding that feature type):

\begin{figure}
\footnotesize
\begin{tabular}{c	@{\hspace{2mm}}| @{\hspace{2mm}}c}
\begin{tabular}{rrp{40mm}}
\smallskip
(a)	& Clinton:	& They're doing it to try to influence the election for \textbf{Donald Trump}.	\\\\
\end{tabular}

& 

\begin{tabular}{rrp{70mm}}
\smallskip
(b)	& Clinton:	& In the days after the first debate, you sent out a series of tweets from 3 a.m.\ to 5 a.m., including one that told people to check out a sex tape.  \\
  & Clinton:	&  Is that the discipline of a good leader? \\
  & Trump		&	\textbf{No, there wasn't check out a sex tape.}
\end{tabular}
\end{tabular}
\caption{Examples of the impact of context in debate instances.}
\label{fig:examples_debate}
\end{figure}

\paragraph{\textbf{Metadata - using opponent's name.}} According to our explanation in Section~\ref{check_worthy_Modeling_Context}, the target sentence in Figure~\ref{fig:examples_debate}(a) mentions the name of the opponent, and this turned out to be the critical feature for correctly predicting that these are check-worthy claims.



\paragraph{\textbf{Contradiction.}} Sometimes, an important claim contains a contradiction to what has been said earlier, e.g., the bold sentence in Figure~\ref{fig:examples_debate}(b). We model the contradiction as explained in Section~\ref{check_worthy_Modeling_Context} to extract such check-worthy claims.


\paragraph{\textbf{Similarity of the sentence to known positive/negative examples.}} The sentence \emph{``For the last seven-and-a-half years, we've seen America's place in the world weakened.''} is similar to the already fact-checked sentence \emph{``We've weakened America's place in the world.''} Thus, the latter is to be classified as check-worthy.

Following, there are some examples for the cQA fact-checking task, where the use of particular contextual features allowed the system to predict correctly the factuality of the answers (they were all misclassified when the corresponding contextual feature was turned off):

\paragraph{\textbf{Support from the current thread.}} The example in Figure~\ref{fig:examples_cqa}(a) shows how the thread information (e.g., similarity of one answer to the other answers in the thread) helps to predict the answer's factuality. The question has four answers that should all be \textsc{True}, but they had been misclassified without the support from the current thread.

\begin{figure}
\footnotesize
\begin{tabular}{c|c}
\begin{tabular}{rrp{58mm}}
\smallskip
(a)	& $q$:	& \emph{what is qtel customer service number?} \\
	& $a_1$:	&  Call 111 \ldots\ and hold the line for at least 30 minutes before being answered. Enjoy Q-tel music. \\
	& $a_2$:	&  call 111 \\
	& $a_3$:	&  111 - Customer service \\
	& $a_4$:	&  111\\
\end{tabular}

&

\begin{tabular}{rrp{72mm}}
\smallskip
(b)	& $q$:	& 
I have visiting visa for work; so can I drive? I have egyptian license \\
	& $a$: & If you are on a visiting Visa and and you have an international driver license you can use it for 6 month I guess. \\\\
	& & \textbf{Evidence}: \emph{[\ldots] A valid international driving license can be used from the day you arrive into the country until six months. [\ldots]}
\end{tabular}
\\\\\hline

\begin{tabular}{rrp{58mm}}

(c)	& $q$: & Smoke after dark during ramadan? \\
	& $a$: & Yes! You can smoke in public after sunset till dawn. \\\\
	& &	\textbf{Evidence}: \emph{Bars and pubs will generally remain open but will only serve alcohol after dark. [\ldots] Don't smoke, drink, chew gum or eat in public during the hours of sunrise to sunset.} \\\\
\end{tabular}

&

\begin{tabular}{rrp{72mm}}
\\\smallskip
(d)	& $q$: & I am in the process of coming over to work in Doha. I wanted to know if their were any specific shots and vaccinations I should get before coming over. I want to try and get these things completed before I leave the US. \\
	& $a$: & Yes there are; though it varies depending on which country you come from. In the UK; the doctor has a list of all countries and the vaccinations needed for each. I'll imagine they have the same in the US.
\end{tabular}
\end{tabular}
 
\caption{Examples of the impact of context and discourse in cQA instances.}
\label{fig:examples_cqa}
\end{figure}


\paragraph{\textbf{Support from high-quality posts in Qatar Living.}} The example in Figure~\ref{fig:examples_cqa}(b) was correctly classified as \textsc{True} when using the high-quality posts, and misclassified as \textsc{False} otherwise. The high-quality posts in the QL forum contain verified information about common topics discussed by people living in Qatar such as visas, driving regulations, customs, etc. The example shows one piece of relevant evidence selected by our method from the high-quality posts, which possibly helps in making the right classification. 


\paragraph{\textbf{Support from all of Qatar Living}} The example in Figure~\ref{fig:examples_cqa}(c) shows the evidence found in the search results in the entire Qatar Living forum. It was classified correctly as \textsc{True} when using the support from all of the Qatar Living forum, and it was misclassified without it.  

%

\subsection{Impact of Discourse}

As the evaluation results have shown, discourse also played an important role.
Let us take the check-worthiness task as an example. In the sentence \emph{``But what President Clinton did, he was impeached, he lost his license to practice law.''}, the discourse parser identified the fragment ``\emph{But what President Clinton did}'' as \textsc{Background} referring to the text for facilitating understanding; the segment ``\emph{he was impeached}'' is \textsc{Elaboration} referring to additional information and ``\emph{$\dots$ to practice law}'' is \textsc{Enablement} referring to the action. These relations are associated with factually-true claims.

Similarly, for cQA fact-checking using discourse information yielded correct classification as \textsc{True} for the example in Figure~\ref{fig:examples_cqa}(d). The question and the answer were parsed together and the segment containing the answer was identified as \textsc{Elaboration}.
The answer further contains a \textsc{Background} segment (``\emph{In the UK; the doctor has a list of all countries and the vaccinations needed for each.}'') and an \textsc{Attribution} segment (``\emph{they have the same in the US}''). These discourse relations are also associated with factually-true answers (as we have seen also in the Figure~\ref{fig:examples_cqa}(c)).

%

\section{Related Work}
\label{sec:related}

Journalists, web users, and researchers are aware of the proliferation of false information on the Web, and as a result, topics such as information credibility and fact-checking are becoming increasingly important as research directions \cite{Lazer1094,Vosoughi1146}. 
For instance, there was a recent special issue of the ACM Transactions on Information Systems journal on Trust and Veracity of Information in Social Media~\cite{Papadopoulos:2016:OSI}, there was a SemEval-2017 shared task on Rumor Detection~\cite{derczynski-EtAl:2017:SemEval}, and there was a lab at CLEF-2018 on Automatic Identification and Verification of Claims in Political Debates \cite{clef2018checkthat:overall,clef2018checkthat:task1,clef2018checkthat:task2}.

\subsection{Detecting Check-Worthy Claims}

The task of detecting check-worthy claims has received relatively little research attention so far.
\citet{Hassan:15} developed \emph{ClaimBuster}, which assigns each sentence in a document a score, i.e.,~a number between 0 and 1 showing how worthy it is for fact-checking. The system is trained on their own dataset of about 8,000 debate sentences (1,673 of them check-worthy), annotated by students, university professors, and journalists. Unfortunately, this dataset is not publicly available, and it contains sentences without context as about 60\% of the original sentences had to be thrown away due to lack of agreement.
In contrast, we developed a new publicly-available dataset based on manual annotations of political debates by nine highly-reputed fact-checking sources, where sentences are annotated in the context of the entire debate. This allows us to explore a novel approach, which focuses on the context.
Note also that the \emph{ClaimBuster} dataset is annotated following guidelines from~\cite{Hassan:15} rather than trying to mimic 
a real fact-checking website; yet, it was later evaluated against 
PolitiFact \cite{Hassan2016ComparingAF}. In contrast, we train and evaluate directly on annotations from fact-checking websites, and thus we learn to fit them better.\footnote{Our model is released as an online demo that supports both English and Arabic \cite{NAACL2018:claimrank}: \url{http://claimrank.qcri.org/}}

\citet{Patwari:17} also focused on the 2016 US Presidential election campaign and independently obtained their data following a similar approach. Their setup asked to predict whether any of the fact-checking sources would select the target sentence. They used a boosting-like model that takes SVMs focusing on different clusters of the dataset and the final outcome was that coming from the most confident classifier. The features considered go from LDA topic-modeling to POS tuples and bag-of-word representations. Unlike that work, we further mimic the selection strategy of one particular fact-checking organization by learning to jointly predict the selection choices by various such organizations. 

The above-mentioned lab on fact-checking at CLEF-2018, was partially based on a variant of our data, but it focused on one fact-checking organization only \cite{clef2018checkthat:task1}, unlike our multi-source setup here.

Beyond the document context, it has been proposed to mine check-worthy claims on the Web. For example, \citet{ennals2010disputed} searched for linguistic cues of disagreement between the author of a statement and what is believed, e.g.,~``falsely claimed that X''. 
The claims matching the patterns go through a statistical classifier, which marks the text of the claim. This procedure can be used to acquire a corpus of disputed claims from the Web. 
Given a set of disputed claims, \citet{ennals2010highlighting} approached the task as locating new claims on the Web that entail the ones that have already been collected. Thus, the task can be conformed as recognizing textual entailment, which is analyzed in detail in \cite{dagan2009recognizing}.
Finally, \citet{le2016towards} argued that the top terms in claim vs.\ non-claim sentences are highly overlapping, which is a problem for bag-of-words approaches. Thus, they used a CNN, where each word is represented by its embedding and each named entity is replaced by its tag, e.g., \emph{person}, \emph{organization}, \emph{location}.

\subsection{Fact-Checking and Credibility}

The credibility of contents on the Web has been questioned by researches for a long time. 
While in the early days the main research focus was on online news portals~\cite{brill2001online,Hardalov2016}, the interest has eventually shifted towards social media \cite{Castillo:2011:ICT:1963405.1963500,PlosONE:2016,popat2017,RANLP2017:clickbait,Vosoughi1146},
which are abundant in sophisticated malicious users such as opinion manipulation \emph{trolls} \cite{InternetResearchJournal:2018} --- paid \cite{Mihaylov2015ExposingPO} or just perceived \cite{Mihaylov2015FindingOM,mihaylov-nakov:2016:P16-2} ---, \emph{sockpuppets} \cite{Maity:2017:DSS:3022198.3026360}, \emph{Internet water army} \cite{Chen:2013:BIW:2492517.2492637}, and \emph{seminar users} \cite{SeminarUsers2017}.

Most of the efforts on assessing credibility have focused on micro-blogging websites. 
For instance, \citet{Canini:2011} studied the credibility of Twitter accounts (as opposed to tweet posts), and found that both the topical content of information sources and social network structure affect source credibility.
Another work, closer to ours, aims at addressing credibility assessment of rumors on Twitter as a problem of finding false information about a newsworthy event~\cite{Castillo:2011:ICT:1963405.1963500}. Their model considered a variety of features including user reputation, writing style, and various time-based features, among others.

Other efforts have focused on news communities. For example, several truth discovery algorithms were studied and combined in an ensemble method for veracity estimation in the VERA system~\cite{Ba:2016:VERA}. They proposed a platform for end-to-end truth discovery from the Web: extracting unstructured information from multiple sources, combining information about single claims, running an ensemble of algorithms, and visualizing and explaining the results.
They also explored two different real-world application scenarios for their system:
fact-checking for crisis situations and evaluation of trustworthiness of a rumor. However, the input to their model is structured data, while here we are interested in unstructured text.
Similarly, the task defined in \cite{mukherjee2015leveraging} combines three objectives: assessing the credibility of a set of posted articles, estimating the trustworthiness of sources, and predicting user's expertise.
They considered a manifold of features characterizing language, topics and Web-specific statistics (e.g., review ratings) on top of a continuous conditional random fields model.
In follow-up work, \citet{popat2016credibility} proposed a model to support or refute claims from \url{snopes.com} and the Wikipedia 
by considering supporting information gathered from the Web. 
In another follow-up work, \cite{popat2017}  proposed a complex model that considers stance, source reliability, language style, and temporal information.

Another important research direction is on using tweets and temporal information for checking the factuality of rumors. For example, \citet{Ma:2015:DRU} used temporal patterns of rumor dynamics to detect false rumors and to predict their frequency. They focused on detecting false rumors in Twitter using time series. They used the change of social context features over a rumor's life cycle in order to detect rumors at an early stage after they were broadcast.

A more general approach for detecting rumors is explored in~\cite{ma2016detecting}, who used recurrent neural networks to learn hidden representations that capture the variation of contextual information of relevant posts over time. Unlike this work, we do not use microblogs, but we query the Web directly in search for evidence.


In the context of question answering, there has been work on assessing the credibility of an answer, e.g., based on intrinsic information, i.e.\ without any external resources~\cite{banerjee-han:2009:NAACLHLT09-Short}. In this case, the reliability of an answer is measured by computing the divergence between language models of the question and of the answer.
The spawn of community-based question answering Websites also allowed for the use of other kinds of information. Click counts, link analysis (e.g., PageRank), and user votes have been used to assess the quality of a posted answer~\cite{Agichtein:2008:FHC:1341531.1341557,Jeon:2006:FPQ:1148170.1148212,Jurczyk:2007:DAQ:1321440.1321575}. Nevertheless, these studies address the answers' credibility level just marginally.

\noindent Efforts to estimate the credibility of an answer in order to assess its overall quality required the inclusion of content-based information~\cite{Su-EtAl:2010:PACLIC2010}, e.g., verbs and adjectives such as \emph{suppose} and \emph{probably}, which cast doubt on the answer. 
Similarly, \citet{lita2005qualitative} used source credibility (e.g., does the document come from a government Website?), sentiment analysis, and answer contradiction compared to other related answers. Another way to assess the credibility of an answer is to incorporate textual entailment methods to find out whether a text (question) can be derived from a hypothesis (answer). 
%
Overall, the \emph{credibility} assessment for question answering has been mostly modeled at the feature level, with the goal of assessing the quality of the answers. A notable exception is the work of~\citet{RANLP2017:credibility:trolls}, where credibility is treated as a task of its own right.
Yet, \emph{credibility} is different from \emph{factuality} (our focus here) as the former is a subjective perception about whether a statement is credible, rather than verifying it as true or false;
still, these notions are often wrongly mixed in the literature.
To the best of our knowledge, no previous work has targeted fact-checking of answers in the context of community Question Answering by gathering external support.


\section{Conclusion and Future Work}
\label{sec:conclusions}



We have studied the role of context and discourse information for two factuality tasks: \Ni detecting check-worthy claims in political debates, and \Nii fact-checking answers in a community question answering forum. We have developed annotated resources for both tasks, which we have made publicly available, and we have proposed rich input representations ---including discourse and contextual features---, and also a complementary set of core features to make our systems as strong as possible. The definition of context varies between the two tasks. For check-worthiness estimation, a target sentence occurs in the context of a political debate, where we model the current intervention by a debate participant in relationship to the previous and to the following participants' turns, together with meta information about the participants, about the reaction of the debate's public, etc. In the answer's factuality checking task, the context for the answer involves the full question-answering thread, the related threads in the entire forum, or the set of related high-quality posts in the forum. 

We trained classifiers for both tasks using neural networks, kernel-based support vector machines, and combinations thereof, and we ran a rigorous evaluation, comparing against alternative systems whenever possible. We also discussed several cases from the test set where the contextual information helped make the right decisions.
Overall, our experimental results and the posterior manual analysis have shown that discourse cues, and especially modeling the context, play an important role and thus should be taken into account when developing models for these tasks.

In future work, we plan to study the role of context and discourse for other related tasks, e.g., for checking the factuality of general claims (not just answers to questions), and for stance classification in the context of factuality. We also plan to experiment with a joint model for check-worthiness estimation, for stance classification, and for fact-checking, which would be useful in an end-to-end system \cite{baly-EtAl:2018:N18-2,NAACL2018:stance}.

We would also like to extend our datasets (e.g., with additional debates, but also with interviews and general discussions), thus enabling better exploitation of deep learning. Especially for the answer verification task, we would like to try distant supervision based on known facts, e.g.,~from high-quality posts, which would allow us to use more training data. We also want to improve user modeling, e.g.,~by predicting factuality for the user's answers and then building a user profile based on that. Finally, we want to explore the possibility of providing justifications for the verified answers, and ultimately of integrating our system in a real-world application.



\bibliographystyle{ACM-Reference-Format}
\bibliography{bibliography}

\end{document}